\documentclass[letterpaper]{article} 
\usepackage[preprint]{aaai2027}  
\usepackage[hyphens]{url}  
\usepackage{graphicx} 
\usepackage{natbib}  
\usepackage{caption} 
\usepackage{algorithm}
\usepackage{algorithmic}

\usepackage{newfloat}
\usepackage{amsmath}
\usepackage{listings}
\DeclareCaptionStyle{ruled}{labelfont=normalfont,labelsep=colon,strut=off} 
\floatstyle{ruled}
\newfloat{listing}{tb}{lst}{}
\floatname{listing}{Listing}

\usepackage{booktabs}
\usepackage{multirow}

\title{FETERS: Few-Shot Early Time-Series Classification via Effective Ratio Selection}

\author{
Chen-An Tai\textsuperscript{1},
Yujia Wu\textsuperscript{1},
Vincent S. Tseng\textsuperscript{1\textdagger}
}

\affiliations{
\textsuperscript{1}Department of Computer Science,
National Yang Ming Chiao Tung University, Taiwan \\
\textsuperscript{\textdagger}Corresponding author: vtseng@cs.nycu.edu.tw
}

\begin{document}

\maketitle

\begin{abstract}
Early time-series classification (ETSC) aims to make accurate predictions from partially observed time series as early as possible. Although various stopping mechanisms and feature learning strategies have been developed for ETSC, most existing methods assume access to sufficient labeled training data, which may be unrealistic in applications with limited annotation. Under limited supervision, learning an additional sample-level stopping module and extracting effective classification features can both become challenging. In this paper, we propose FETERS, a few-shot ETSC framework that selects a dataset-level stopping ratio through class-wise leave-one-out (LOO) evaluation on the support set and uses a penalty-based reward function to manage the accuracy--earliness trade-off, thereby avoiding the need to train an additional stopping module. FETERS further combines Rocket-based features with frozen Chronos representations for classification. Extensive experiments on 69 public datasets spanning 14 domains show that FETERS achieves state-of-the-art (SOTA) performance in the 5-shot setting, with the highest average harmonic mean (HM) and the best HM on 38 datasets, while outperforming the current SOTA method on 44 datasets. FETERS also remains competitive in the full-shot setting, demonstrating its effectiveness in managing the accuracy--earliness trade-off.
\end{abstract}


\section{Introduction}

\begin{figure}[!t]
\centering
\includegraphics[width=0.9\columnwidth]{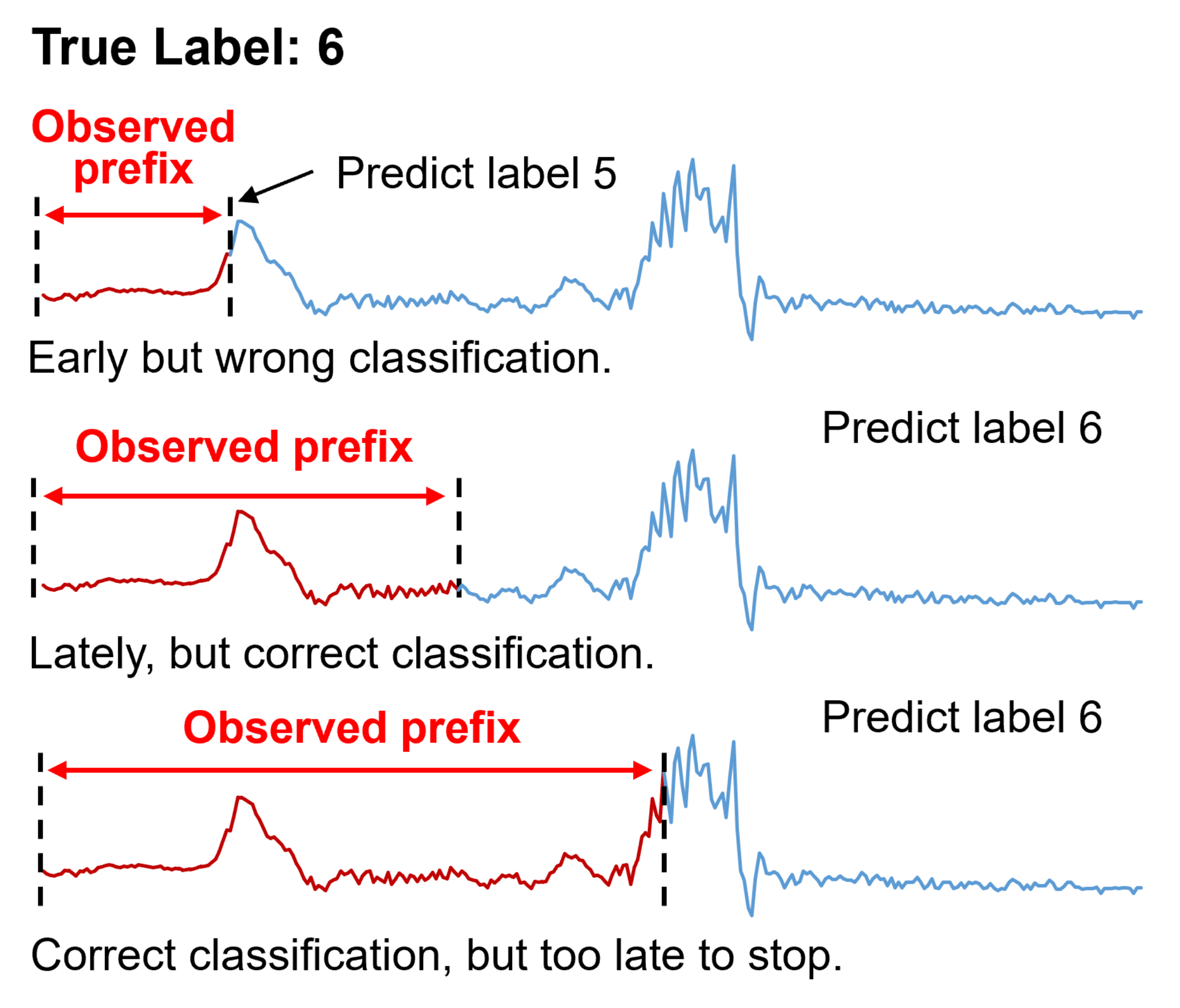}
\caption{Illustration of the accuracy--earliness trade-off in ETSC using a sample from the UCR Archive~\cite{UCRArchive2018}. Shorter observed prefixes yield earlier predictions and greater earliness.}
\label{fig:tradeoff}
\end{figure}

Time series classification (TSC) aims to classify observations collected over time~\cite{classification_review}. Early time series classification (ETSC) further seeks to make predictions from partial observations as early as possible while preserving classification accuracy. Figure~\ref{fig:tradeoff} illustrates this accuracy--earliness trade-off. Existing methods commonly determine when to predict at the sample level using learned halting policies, confidence criteria, or estimates of future classification costs~\cite{renault2025earlyclassificationtimeseries}.

ETSC is relevant to time-sensitive domains such as healthcare, industrial monitoring, and human activity recognition~\cite{gupta2020approaches}, where labeled time series data may be limited by costly annotation, specialized data collection, or rare target events~\cite{Liang2023industryfew, shyalika2024fewrole}. Although few-shot learning has received increasing attention in these domains~\cite{serpush2022wear, wang2025few, li2025fewshot}, recent studies have primarily focused on conventional TSC rather than ETSC~\cite{liu2025novel, vettoruzzo2025efficient}. Extending ETSC to few-shot supervision introduces two challenges: First, modeling an additional sample-level stopping mechanism from a small support set increases the complexity of the learning problem and may make the resulting decisions sensitive to the sampled examples. Second, time-series datasets from different domains can exhibit substantially different temporal patterns, making it challenging to extract effective representations from partially observed prefixes.

To the best of our knowledge, ETSC under per-class few-shot supervision has not yet been systematically studied. Existing methods are primarily developed under full-data settings, leaving their behavior under limited supervision unclear. To investigate this setting, we propose FETERS, a few-shot ETSC framework that trades sample-level stopping flexibility for a simpler dataset-level ratio-selection problem. FETERS estimates candidate observation ratios through class-wise leave-one-out (LOO) evaluation on the support set and applies the selected ratio to all test samples from the same dataset. This formulation avoids estimating an additional stopping module from limited labeled data while retaining adaptation to dataset-specific temporal characteristics. FETERS further combines dataset-fitted Rocket features with frozen Chronos representations for prefix classification.

The main contributions of this work are summarized as follows:
\begin{itemize}
    \item To the best of our knowledge, we present the first systematic study of ETSC under per-class few-shot supervision and evaluate representative ETSC methods across multiple shot settings.
    \item We propose a dataset-level ratio selection strategy that evaluates candidate observation ratios through support-set class-wise LOO evaluation and uses a penalty parameter to control the preferred accuracy--earliness trade-off, without requiring an additional learned stopping module.
    \item We combine dataset-fitted Rocket features with frozen Chronos representations for few-shot prefix classification. Extensive experiments on 69 datasets across 14 domains show that FETERS achieves the best average HM and HM rank among the evaluated methods in the 5-shot setting, while remaining competitive under full-shot supervision.
\end{itemize}

\section{Related Work}

\subsection{Stopping Mechanisms}
Existing stopping mechanisms can be broadly categorized as confidence-based, agent-based, and anticipation-based methods~\cite{renault2025earlyclassificationtimeseries}. Confidence-based methods stop according to the confidence of current predictions~\cite{Lv2019ECEC, schafer2020teaser, Lv2023secondorder, Yan2025UTSC}, whereas agent-based methods learn whether to halt or continue as observations arrive~\cite{har2019earliest, har2022stopandhop, huang2024SPN}. Anticipation-based methods use non-myopic criteria that account for expected future predictions or the cost of waiting~\cite{Tavenard2016Cost, achenchabe2021Economy, jakub2023CALIMERA}. 

These methods are primarily developed under full-data supervision and generally make sample-level stopping decisions. FETERS instead selects a dataset-level observation ratio through support-set evaluation and applies it to all test samples from the same dataset. This formulation sacrifices sample-level flexibility but avoids learning an additional stopping module from a small support set.

\subsection{Feature Extraction}
Both non-deep and deep feature extractors have been used in ETSC. Non-deep approaches include WEASEL~\cite{Lv2019ECEC, schafer2020teaser} and MiniRocket-based pipelines~\cite{jakub2023CALIMERA, Yan2025UTSC}. WEASEL represents time-series subsequences as discrete words using Fourier-based representations~\cite{sch2017WEASEL}, whereas MiniRocket extracts convolution features using a largely fixed set of kernels with different dilations and biases~\cite{Dempster2021minirocket}. Deep approaches use architectures such as FCN-Transformer combinations~\cite{Lv2023secondorder}, directional LSTMs~\cite{Marc2023corp}, and CNN-LSTM models~\cite{huang2024SPN} to learn representations from partial observations. Although these methods have proven effective in ETSC, they were primarily developed in full-shot settings. Learning high-capacity representations directly from a small support set may be particularly challenging, especially when classification must be performed from partially observed prefixes~\cite{Wen2021deeplsurvey}.

Time-series foundation models provide another source of representations learned from large external corpora. Chronos is pretrained for probabilistic time-series forecasting and has demonstrated zero-shot generalization to unseen forecasting datasets~\cite{ansari2024chronos}. This suggests that forecasting-pretrained models may capture general temporal information. Recent studies have further explored the use of time-series foundation models in TSC~\cite{Zhuang2025VectorICL, liu2026unified}. FETERS therefore combines dataset-fitted MiniRocket/MultiRocket~\cite{tan2022multirocket} features with frozen Chronos representations and empirically evaluates the individual and combined contributions of these feature sources under both few-shot and full-shot supervision.

\section{Proposed Method}
In this section, we introduce FETERS, a few-shot ETSC framework comprising two main components: ratio selection and hybrid feature extraction, which jointly manage the accuracy--earliness trade-off in few-shot ETSC.

\begin{figure*}
\centering
\includegraphics[width=0.95\textwidth]{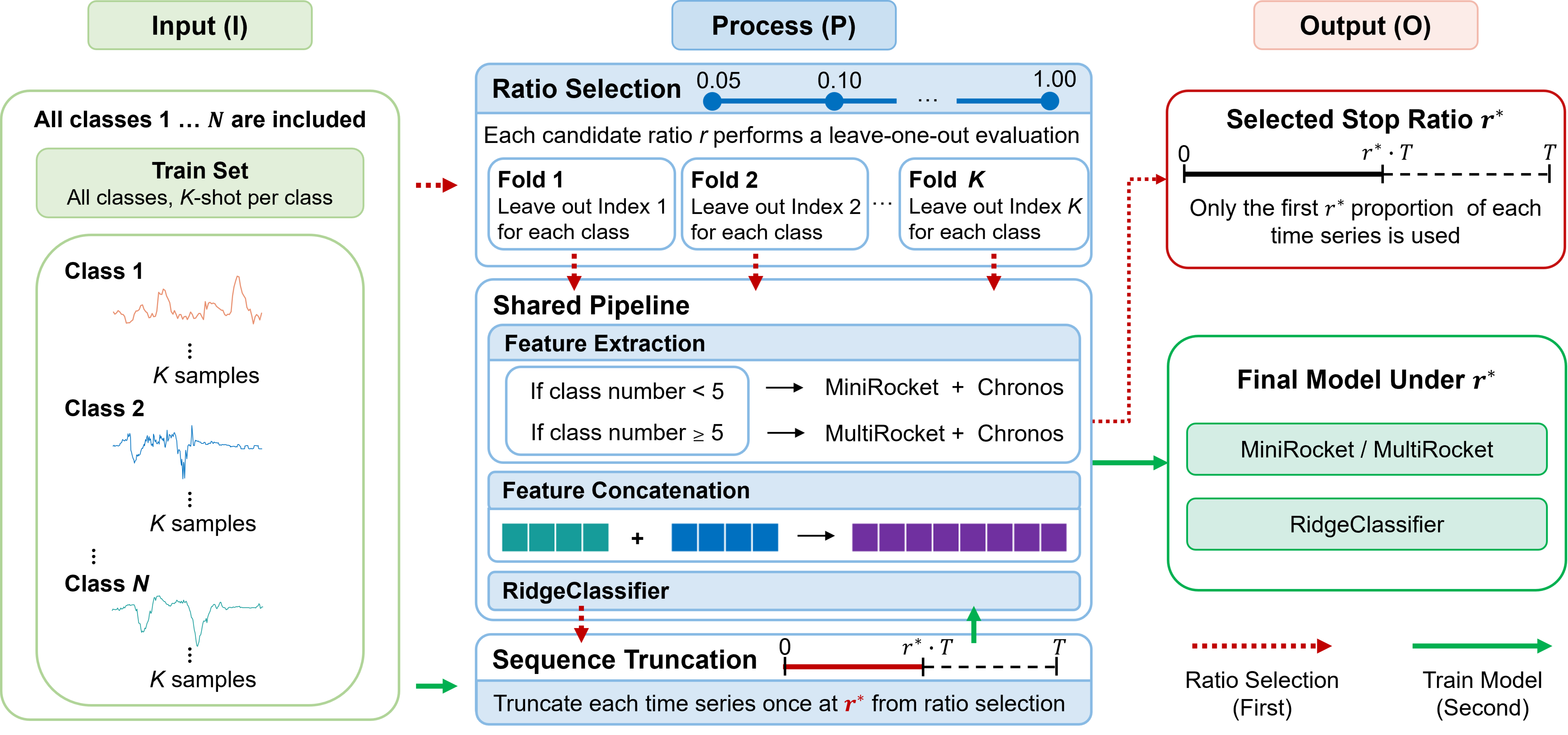}
\caption{Overview of the FETERS framework. The figure illustrates the training workflow, where the red path first performs ratio selection and the green path then trains the shared pipeline using the selected ratio $r^{*}$. During inference, test samples are truncated using $r^{*}$ and processed by the trained model following the same shared pipeline.}
\label{fig:framework}
\end{figure*}

\subsection{Problem Description}

For each time-series dataset, let $\mathcal{N}$ denote the set of classes. For each class $c \in \mathcal{N}$, $K$ labeled samples are selected to form the class-specific support set $S_c=\{(x_c^{(1)},y_c^{(1)}),\ldots,(x_c^{(K)},y_c^{(K)})\}$, where $x_c^{(k)}$ denotes the $k$-th time-series sample from class $c$ and $y_c^{(k)}=c$. The complete support set is defined as $S=\bigcup_{c\in\mathcal{N}} S_c$. Let $\mathcal{R}=\{0.05,0.10,0.15,\ldots,1.00\}$ denote the set of candidate observation ratios. For a time-series sample $x$ of length $T$, the prefix observed at ratio $r\in\mathcal{R}$ is defined as $x_{1:\max(2,\lfloor rT\rfloor)}$, which is used for feature extraction and classification at ratio $r$. The goal is to select a dataset-level observation ratio that balances classification accuracy and earliness.

\subsection{Ratio Selection}
Under limited labeled data, modeling an additional sample-level stopping module increases the complexity of the learning problem and may make stopping decisions sensitive to the sampled support examples. FETERS instead adopts a dataset-level ratio selection strategy based on class-wise LOO evaluation over the support set. For each candidate ratio $r \in \mathcal{R}$, FETERS evaluates the shared pipeline shown in Figure~\ref{fig:framework} on the support set through leave-one-out folds. The number of leave-one-out folds is determined by the number of shots $K$, with a maximum of $20$ folds to control the computational cost in full-shot settings. In each fold, the samples with the same index $i$ are held out from every class for evaluation whenever available, while the remaining support samples are used for training, where $i=1,\ldots,\min(K,20)$. The classification accuracy is then averaged over all folds to estimate the performance of the ratio $r$.

\begin{algorithm}[!t]
\caption{Ratio Selection Algorithm}
\label{alg:ratio_selection}
\textbf{Input}: Support set $S$, shot number $K$, number of classes $N$\\
\textbf{Parameter}: Candidate ratios $\mathcal{R}$, $M_{\max}=20$, $p$\\
\textbf{Output}: Selected stopping ratio $r^*$
\begin{algorithmic}[1]
\STATE $M \leftarrow \min(K,M_{\max})$
\FOR{$r \in \mathcal{R}$}
    \STATE $\mathcal{A}_r \leftarrow \emptyset$
    \FOR{$i=1,\ldots,M$}
        \STATE $S_{\mathrm{val}}^{i} \leftarrow \emptyset$
        \FOR{$n=1,\ldots,N$}
            \STATE $S_{\mathrm{val}}^{i} \leftarrow S_{\mathrm{val}}^{i} \cup \{(x_{n}^{i}, n)\}$
        \ENDFOR
        \STATE $S_{\mathrm{train}}^{i} \leftarrow S \setminus S_{\mathrm{val}}^{i}$
        \STATE $S_{\mathrm{train}}^{i,r}, S_{\mathrm{val}}^{i,r} \leftarrow \mathrm{Truncate}(S_{\mathrm{train}}^{i}, S_{\mathrm{val}}^{i}, r)$
        \STATE Fit $P_{r,i}$ on $S_{\mathrm{train}}^{i,r}$
        \STATE $a_{r,i} \leftarrow \mathrm{Acc}(P_{r,i}, S_{\mathrm{val}}^{i,r})$
        \STATE $\mathcal{A}_r \leftarrow \mathcal{A}_r \cup \{a_{r,i}\}$
    \ENDFOR
    \STATE $\mathrm{Acc}(r) \leftarrow \mathrm{mean}(\mathcal{A}_r)$
    \STATE $\mathrm{Reward}(r) \leftarrow \mathrm{Acc}(r)(1- 0.95r^p)$
\ENDFOR
\STATE $r^* \leftarrow \arg\max_{r\in\mathcal{R}}\mathrm{Reward}(r)$
\STATE \textbf{return} $r^*$
\end{algorithmic}
\end{algorithm}

Since accuracy alone does not account for earliness, which is a key objective in ETSC, we further apply a penalty-based reward function Eq.~\ref{eq:ratio_selection_reward} to model the accuracy--earliness trade-off:
\begin{equation}
\begin{gathered}
\mathrm{Reward}(r) = \mathrm{Acc}(r)\left(1-0.95r^p\right), \\
r^* = \arg\max_{r\in\mathcal{R}} \mathrm{Reward}(r),
\end{gathered}
\label{eq:ratio_selection_reward}
\end{equation}
where $\mathrm{Acc}(r)$ denotes the classification accuracy averaged across support-set folds, and $p$ controls how strongly later decisions are penalized. Smaller values of $p$ impose a stronger relative penalty at early and intermediate ratios, whereas larger values reduce the penalty on delayed decisions and therefore tend to favor later predictions with higher accuracy. The coefficient $0.95$ keeps the penalty factor positive at $r=1.0$.

Algorithm~\ref{alg:ratio_selection} summarizes the ratio selection procedure, where $P_{r,i}$ denotes the shared pipeline and $\mathrm{Truncate}(\cdot,\cdot,r)$ denotes prefix truncation at ratio $r$. The selected ratio $r^*$ is then used to truncate the support samples for training the final classifier and the test samples for inference, thereby controlling the accuracy--earliness trade-off for the dataset.

\subsection{Feature Extraction}
In ETSC, prefixes observed at different ratios may exhibit distinct temporal characteristics, making effective feature extraction particularly important for accurate classification. FETERS extracts two types of features from each observed prefix: Rocket-based and Chronos-based features. 

Previous ETSC studies have also adopted architectures in which multiple feature extraction modules serve distinct roles rather than relying on a single feature extractor~\cite{Lv2023secondorder, huang2024SPN}. In FETERS, Rocket-based feature extraction provides efficient convolution-based representations, while Chronos-based feature extraction provides additional temporal representations from a time-series foundation model, offering an additional general feature source beyond the Rocket-based extractor. The two feature vectors are then concatenated and used as the final representation for RidgeClassifier.

For Rocket-based feature extraction, we apply a fixed class-count rule across all datasets: MiniRocket~\cite{Dempster2021minirocket} is used when the number of classes is fewer than five, and MultiRocket~\cite{tan2022multirocket} is used otherwise. This rule was selected from an efficiency–effectiveness analysis comparing the two extractors across class-count groups rather than through dataset-specific tuning. The selected extractor is fitted on the truncated support samples and subsequently applied to both support and test prefixes. For Chronos-based feature extraction, we use the pretrained Chronos model~\cite{ansari2024chronos} as a frozen feature encoder. Chronos yields a fixed-dimensional representation for each observed prefix. The Rocket and Chronos representations are concatenated, and RidgeClassifier is fitted on the resulting support-set features at the selected ratio $r^*$, and then used to classify test samples truncated at the same ratio.

\section{Experiments}
\subsection{Experimental Setup}
\textbf{Datasets}\quad We evaluate our method on 69 datasets from the UCR Time Series Archive~\cite{UCRArchive2018}, spanning 14 UCR-defined domains. This includes 45 classical ETSC benchmark datasets widely used in prior studies~\cite{Lv2019ECEC, schafer2020teaser, jakub2023CALIMERA, Lv2023secondorder, Yan2025UTSC}, covering 6 domains: Image, Sensor, Motion, Spectro, ECG, and Simulated. We further include 24 additional datasets from another 8 domains: Device, Hemodynamics, EOG, EPG, Power, Spectrum, Traffic, and Trajectory, allowing us to evaluate our method across a broader range of time-series domains.

\noindent
\textbf{Evaluation Metrics}\quad We evaluate ETSC performance using three metrics: accuracy, earliness, and harmonic mean (HM, Eq.~\ref{eq:HM}). Accuracy measures classification correctness, while earliness measures the proportion of time series observed before a prediction is made; smaller values indicate earlier predictions. HM jointly evaluates accuracy and earliness, where larger values indicate a better trade-off.

\begin{equation}
\begin{gathered}
\mathrm{HM}=
\frac{2\cdot \mathrm{Accuracy}\cdot (1-\mathrm{Earliness})}
{\mathrm{Accuracy}+(1-\mathrm{Earliness})}
\end{gathered}
\label{eq:HM}
\end{equation}

\noindent
\textbf{Baselines} We compare FETERS with four representative ETSC methods, including classical benchmark methods and recent state-of-the-art approaches. \textbf{(1)} EARLIEST \cite{har2019earliest} jointly trains a recurrent classifier and a reinforcement learning controller, with $\lambda$ balancing accuracy and earliness. \textbf{(2)} TEASER~\cite{schafer2020teaser} uses a classifier for early prediction and a one-class SVM to determine whether the prediction is sufficiently reliable based on its confidence. \textbf{(3)} CALIMERA~\cite{jakub2023CALIMERA} uses calibrated probabilities to estimate classification costs and regression to predict whether waiting will reduce those costs. \textbf{(4)} UTSC~\cite{Yan2025UTSC} uses an Uncertainty Probability Decoder and a reweighting mechanism to model uncertainty, with a threshold $\delta$ to determine whether a prediction is sufficiently reliable.

For consistency and reproducibility, we use the available implementations and the hyperparameter settings recommended in the original studies. Under the full-shot setting, each method is evaluated once using the configuration recommended in its original study. In few-shot settings, the same fixed configurations are applied without dataset-specific retuning, and all methods use the same random seeds. In the 5-shot setting, all methods are evaluated over 100 runs with seeds ranging from 40 to 139, and the results are averaged across runs.

\begin{table}[t]
\centering
\small
\setlength{\tabcolsep}{1mm}
\begin{tabular}{l|ccc|ccc}
\hline
\multirow{2}{*}{Method}
& \multicolumn{3}{c|}{5-shot} 
& \multicolumn{3}{c}{Full-shot} \\
\cline{2-7}
& Acc. $\uparrow$ & Ear. $\downarrow$ & Wins
& Acc. $\uparrow$ & Ear. $\downarrow$ & Wins \\
\hline
EARLIEST & 0.289 & \textbf{0.018} & 0 & 0.341 & \textbf{0.063} & 1 \\
TEASER   & 0.599 & 0.305 & \textbf{26} & 0.671 & 0.212 & 13\\
UTSC     & 0.611 & 0.296 & 15 & 0.701 & 0.239 & 18\\
CALIMERA & 0.536 & 0.113 & 4 & 0.702 & 0.171 & \textbf{22}\\
FETERS (Ours)   & \textbf{0.628} & 0.213 & \textbf{26} & \textbf{0.725} & 0.197 & \textbf{22}\\
\hline
\end{tabular}
\caption{Average accuracy, earliness, and number of datasets with the highest accuracy in the 5-shot and full-shot settings.}
\label{tab:avg_performance}
\end{table}

\subsection{Overall Performance Comparison}
\noindent
\textbf{Accuracy and Earliness}\quad As shown in Table~\ref{tab:avg_performance}, FETERS achieves the highest average accuracy in both settings, reaching 62.8\% in the 5-shot setting and 72.5\% in the full-shot setting, and records 26 and 22 dataset-level accuracy wins, respectively.

\begin{table*}[!t]
\centering
\small
\renewcommand{\arraystretch}{0.7}
\setlength{\tabcolsep}{1.2mm}
\begin{tabular}{@{}l|ccccc|ccccc@{}}
\hline
\multirow{2}{*}{Dataset}
& \multicolumn{5}{c|}{5-shot HM}
& \multicolumn{5}{c}{Full-shot HM} \\
\cline{2-11}
& EARLIEST & TEASER & UTSC & CALIMERA & FETERS
& EARLIEST & TEASER & UTSC & CALIMERA & FETERS \\
\hline
FiftyWords & 0.069 & 0.555 & 0.532 & 0.476 & \textbf{0.571}
& 0.241 & 0.661 & 0.667 & 0.668 & \textbf{0.671} \\
Adiac & 0.128 & 0.724 & 0.728 & \textbf{0.747} & 0.743 & 0.064 & 0.745 & 0.753 & \textbf{0.771} & 0.748 \\
Beef & 0.360 & 0.657 & 0.627 & 0.787 & \textbf{0.873} & 0.331 & 0.709 & 0.664 & 0.805 & \textbf{0.925} \\
CBF & 0.545 & 0.231 & 0.788 & \textbf{0.811} & 0.808 & 0.495 & 0.645 & 0.809 & \textbf{0.826} & 0.814 \\
ChlorineConc. & 0.471 & 0.329 & 0.340 & \textbf{0.508} & 0.499 & 0.699 & 0.688 & 0.728 & 0.718 & \textbf{0.737} \\
CinCECGTorso & 0.404 & \textbf{0.784} & 0.735 & 0.727 & 0.736 & 0.409 & \textbf{0.859} & 0.792 & 0.814 & 0.790 \\
Coffee & 0.641 & 0.719 & 0.840 & 0.841 & \textbf{0.847} & 0.461 & 0.807 & \textbf{0.930} & 0.929 & 0.889 \\
CricketX & 0.227 & 0.491 & 0.540 & 0.414 & \textbf{0.595} & 0.257 & 0.652 & 0.610 & \textbf{0.699} & 0.675 \\
CricketY & 0.217 & 0.483 & 0.535 & 0.419 & \textbf{0.593} & 0.281 & 0.695 & 0.713 & \textbf{0.716} & 0.715 \\
CricketZ & 0.241 & 0.493 & 0.541 & 0.379 & \textbf{0.606} & 0.254 & 0.672 & 0.680 & \textbf{0.711} & 0.702 \\
DiatomSizeRed. & 0.472 & 0.816 & 0.773 & 0.883 & \textbf{0.884} & 0.460 & 0.856 & 0.839 & 0.892 & \textbf{0.896} \\
ECG200 & 0.666 & 0.696 & 0.744 & \textbf{0.789} & 0.774 & 0.773 & 0.840 & 0.851 & \textbf{0.878} & 0.857 \\
ECGFiveDays & 0.680 & 0.651 & 0.752 & \textbf{0.773} & 0.758 & 0.694 & 0.725 & 0.800 & 0.807 & \textbf{0.826} \\
FaceAll & 0.345 & 0.601 & 0.758 & \textbf{0.766} & 0.728 & 0.411 & 0.814 & 0.792 & \textbf{0.850} & 0.831 \\
FaceFour & 0.364 & 0.680 & \textbf{0.826} & 0.817 & 0.820 & 0.431 & 0.670 & 0.835 & 0.832 & \textbf{0.836} \\
FacesUCR & 0.202 & 0.661 & 0.690 & \textbf{0.692} & 0.679 & 0.300 & \textbf{0.775} & 0.720 & 0.763 & 0.750 \\
Fish & 0.260 & 0.717 & 0.739 & 0.721 & \textbf{0.752} & 0.228 & 0.824 & 0.820 & \textbf{0.859} & 0.825 \\
GunPoint & 0.668 & 0.709 & 0.725 & 0.681 & \textbf{0.754} & 0.651 & 0.800 & 0.780 & \textbf{0.832} & 0.785 \\
Haptics & 0.340 & 0.424 & 0.348 & \textbf{0.445} & 0.437 & 0.360 & 0.500 & 0.525 & 0.571 & \textbf{0.575} \\
InlineSkate & 0.260 & 0.431 & 0.324 & 0.401 & \textbf{0.444} & 0.269 & 0.543 & 0.477 & 0.486 & \textbf{0.550} \\
ItalyPowerDem. & 0.702 & 0.524 & 0.653 & 0.700 & \textbf{0.739} & \textbf{0.760} & 0.721 & 0.710 & 0.714 & 0.745 \\
Lightning2 & \textbf{0.696} & 0.633 & 0.637 & 0.677 & 0.656 & 0.695 & 0.767 & 0.703 & 0.702 & \textbf{0.776} \\
Lightning7 & 0.291 & 0.515 & \textbf{0.516} & 0.424 & 0.510 & 0.383 & 0.567 & 0.608 & 0.545 & \textbf{0.632} \\
Mallat & 0.225 & 0.575 & 0.641 & 0.639 & \textbf{0.644} & 0.220 & 0.607 & 0.660 & \textbf{0.681} & 0.662 \\
MedicalImages & 0.205 & 0.469 & 0.496 & 0.566 & \textbf{0.568} & 0.675 & 0.742 & 0.728 & \textbf{0.775} & 0.769 \\
MoteStrain & 0.792 & 0.739 & 0.813 & 0.834 & \textbf{0.836} & 0.644 & 0.843 & 0.833 & 0.855 & \textbf{0.877} \\
NIFECGThorax1 & 0.093 & \textbf{0.851} & 0.825 & 0.849 & 0.849 & 0.172 & 0.888 & 0.889 & 0.912 & \textbf{0.909} \\
OliveOil & 0.424 & 0.869 & 0.876 & \textbf{0.891} & 0.874 & 0.571 & 0.913 & 0.890 & \textbf{0.917} & 0.875 \\
OSULeaf & 0.296 & 0.569 & 0.559 & 0.536 & \textbf{0.650} & 0.333 & 0.711 & 0.758 & 0.762 & \textbf{0.797} \\
NIFECGThorax2 & 0.133 & 0.882 & 0.868 & \textbf{0.884} & 0.877 & 0.386 & 0.926 & 0.917 & 0.926 & \textbf{0.930} \\
SonyAIBORS1 & 0.809 & 0.734 & 0.754 & 0.867 & \textbf{0.872} & 0.594 & 0.803 & 0.755 & 0.859 & \textbf{0.897} \\
SonyAIBORS2 & 0.715 & 0.720 & 0.771 & 0.797 & \textbf{0.803} & 0.753 & 0.785 & 0.759 & \textbf{0.816} & 0.794 \\
StarLightCurv. & 0.673 & 0.840 & 0.835 & 0.834 & \textbf{0.841} & 0.898 & 0.926 & 0.911 & \textbf{0.939} & 0.917 \\
SwedishLeaf & 0.207 & 0.732 & 0.762 & 0.771 & \textbf{0.775} & 0.331 & 0.824 & 0.825 & \textbf{0.842} & 0.836 \\
Symbols & 0.522 & 0.783 & 0.802 & \textbf{0.814} & 0.792 & 0.505 & 0.794 & 0.804 & \textbf{0.826} & 0.794 \\
SyntheticCtrl. & 0.534 & 0.700 & \textbf{0.768} & 0.754 & 0.734 & 0.719 & \textbf{0.838} & 0.829 & 0.830 & 0.833 \\
Trace & 0.662 & 0.555 & 0.750 & \textbf{0.778} & 0.772 & 0.653 & 0.605 & 0.769 & 0.793 & \textbf{0.820} \\
TwoPatterns & 0.400 & 0.024 & 0.427 & 0.405 & \textbf{0.439} & 0.408 & 0.564 & 0.542 & \textbf{0.580} & 0.405 \\
UWaveGestureX & 0.360 & 0.554 & 0.546 & 0.406 & \textbf{0.572} & 0.419 & 0.638 & 0.699 & \textbf{0.706} & 0.695 \\
UWaveGestureY & 0.376 & \textbf{0.533} & 0.495 & 0.400 & 0.519 & 0.438 & 0.614 & 0.651 & \textbf{0.656} & \textbf{0.656} \\
UWaveGestureZ & 0.373 & 0.537 & 0.514 & 0.427 & \textbf{0.543} & 0.435 & 0.619 & \textbf{0.688} & 0.682 & 0.679 \\
TwoLeadECG & 0.689 & 0.700 & 0.757 & 0.704 & \textbf{0.767} & 0.718 & 0.771 & 0.844 & \textbf{0.847} & 0.813 \\
Wafer & 0.650 & 0.745 & \textbf{0.862} & 0.837 & 0.822 & 0.928 & 0.907 & 0.953 & 0.942 & \textbf{0.973} \\
WordSyn. & 0.120 & \textbf{0.510} & 0.495 & 0.361 & 0.502 & 0.358 & 0.622 & 0.619 & \textbf{0.625} & 0.620 \\
Yoga & 0.660 & 0.645 & 0.633 & \textbf{0.703} & 0.685 & 0.710 & 0.811 & 0.854 & \textbf{0.856} & 0.838 \\
ACSF1 & 0.186 & \textbf{0.770} & 0.721 & 0.739 & 0.740 & 0.225 & 0.793 & 0.771 & 0.804 & \textbf{0.830} \\
Computers & 0.667 & 0.689 & 0.633 & \textbf{0.716} & 0.691 & 0.591 & 0.771 & 0.735 & 0.781 & \textbf{0.785} \\
ElectricDev. & 0.308 & \textbf{0.511} & 0.445 & 0.458 & 0.504 & 0.021 & 0.637 & 0.625 & \textbf{0.654} & 0.605 \\
LargeKitchenApp. & 0.500 & 0.515 & 0.497 & \textbf{0.574} & 0.545 & 0.483 & 0.646 & 0.571 & 0.681 & \textbf{0.682} \\
House20 & 0.678 & 0.764 & 0.799 & \textbf{0.867} & 0.859 & 0.543 & 0.787 & 0.895 & \textbf{0.913} & 0.893 \\
RefrigerationDev. & 0.505 & 0.529 & 0.487 & \textbf{0.562} & \textbf{0.562} & 0.549 & \textbf{0.656} & 0.611 & 0.621 & 0.573 \\
ScreenType & 0.499 & 0.485 & 0.388 & \textbf{0.500} & 0.492 & 0.494 & 0.524 & 0.445 & \textbf{0.568} & 0.551 \\
SmallKitchenApp. & 0.504 & 0.705 & 0.631 & 0.708 & \textbf{0.715} & 0.730 & 0.790 & 0.664 & \textbf{0.841} & 0.834 \\
PigAirwayP. & 0.032 & 0.223 & \textbf{0.425} & 0.232 & \textbf{0.425} & 0.058 & 0.235 & 0.416 & 0.236 & \textbf{0.426} \\
PigArtP. & 0.028 & 0.602 & 0.766 & 0.722 & \textbf{0.796} & 0.049 & 0.397 & \textbf{0.776} & 0.707 & 0.774 \\
PigCVP & 0.042 & 0.478 & 0.584 & 0.243 & \textbf{0.613} & 0.020 & 0.449 & 0.592 & 0.250 & \textbf{0.599} \\
Chinatown & 0.859 & 0.818 & 0.919 & \textbf{0.947} & 0.939 & 0.803 & 0.831 & \textbf{0.956} & 0.950 & 0.950 \\
MelbournePed. & 0.189 & 0.580 & 0.660 & 0.656 & \textbf{0.669} & 0.361 & 0.643 & 0.734 & 0.720 & \textbf{0.761} \\
GestureMidAir1 & 0.271 & 0.492 & 0.502 & 0.512 & \textbf{0.613} & 0.289 & 0.519 & 0.541 & 0.605 & \textbf{0.655} \\
GestureMidAir2 & 0.227 & 0.467 & 0.474 & 0.254 & \textbf{0.560} & 0.255 & 0.445 & 0.452 & 0.476 & \textbf{0.592} \\
GestureMidAir3 & 0.161 & 0.390 & 0.378 & 0.142 & \textbf{0.466} & 0.169 & 0.420 & 0.419 & 0.206 & \textbf{0.519} \\
Rock & 0.483 & 0.609 & 0.739 & 0.733 & \textbf{0.773} & 0.571 & 0.603 & 0.743 & 0.733 & \textbf{0.775} \\
SemgGenderCh2 & 0.686 & 0.676 & 0.695 & 0.713 & \textbf{0.715} & 0.755 & 0.779 & 0.816 & 0.830 & \textbf{0.843} \\
SemgSubjectCh2 & 0.387 & 0.517 & 0.527 & 0.525 & \textbf{0.591} & 0.488 & 0.679 & 0.772 & 0.770 & \textbf{0.796} \\
SemgMoveCh2 & 0.336 & 0.516 & 0.465 & 0.483 & \textbf{0.528} & 0.413 & 0.645 & 0.645 & 0.643 & \textbf{0.673} \\
InsectEPGReg. & 0.496 & 0.837 & \textbf{0.975} & 0.972 & 0.967 & 0.651 & 0.947 & \textbf{0.988} & 0.974 & 0.974 \\
EOGHorizontal & 0.292 & 0.161 & 0.382 & 0.250 & \textbf{0.456} & 0.389 & 0.200 & 0.357 & \textbf{0.477} & 0.429 \\
EOGVertical & 0.276 & 0.147 & 0.360 & 0.216 & \textbf{0.440} & 0.305 & 0.146 & 0.230 & \textbf{0.427} & 0.425 \\
PowerCons & \textbf{0.680} & 0.580 & 0.632 & 0.666 & 0.653 & 0.741 & 0.740 & \textbf{0.827} & 0.798 & 0.796 \\
\cline{1-11}
\rule{0pt}{2.0ex}AVG Mean & 0.412 & 0.592 & 0.638 & 0.628 & \textbf{0.679} & 0.460 & 0.689 & 0.718 & 0.734 & \textbf{0.749} \\
\rule{0pt}{1.0ex}Wins & 2 & 6 & 6 & 19 & \textbf{38} & 1 & 4 & 6 & 29 & \textbf{30} \\
\hline
\end{tabular}
\caption{Per-dataset HM comparison under the 5-shot and full-shot settings. Bold indicates the best result for each dataset and setting. Wins count the best results across datasets. AVG Mean reports the average HM over all datasets.}
\label{tab:hm_5shot_fullshot}
\end{table*}

\noindent
\textbf{HM}\quad As shown in Table~\ref{tab:hm_5shot_fullshot}, FETERS achieves the highest average HM in both settings, reaching 0.679 in the 5-shot setting and 0.749 in the full-shot setting. It achieves the best HM on 38 and 30 of the 69 datasets in the 5-shot and full-shot settings, respectively. Pairwise, FETERS outperforms CALIMERA on 44 datasets in the 5-shot setting, whereas the two methods remain statistically indistinguishable under full-shot supervision. These results indicate that FETERS provides the strongest overall accuracy–earliness trade-off among the evaluated methods under the 5-shot protocol while remaining competitive with established methods when full training data are available.

\begin{figure}[!t]
\centering
\includegraphics[width=0.95\columnwidth]{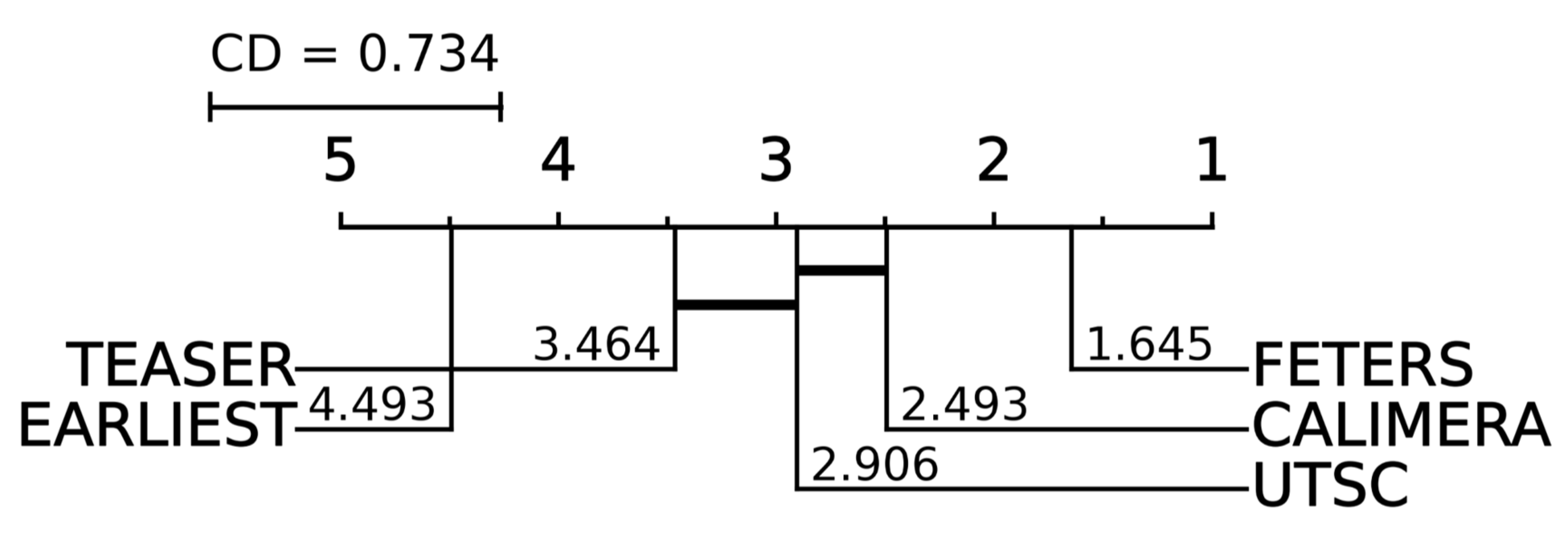}
\caption{Critical difference diagram of HM in 5-shot setting.}
\label{fig:5_shot_HM_statistic}
\end{figure}

\noindent
\textbf{Statistical Significance}\quad Friedman tests followed by post hoc Nemenyi comparisons~\cite{demvsar2006statistical} are used to assess differences in HM ranks across datasets. As shown in Figure~\ref{fig:5_shot_HM_statistic}, FETERS achieves the best average HM rank in the 5-shot setting and is significantly better than all four evaluated baselines at $\alpha=0.05$. Pairwise Wilcoxon signed-rank tests~\cite{wilcoxon1945individual} for with the Holm correction yield the same conclusion for 5-shot HM, including a significant difference against CALIMERA. Under full-shot supervision, however, FETERS and CALIMERA are not significantly different.

\subsection{Stopping-Behavior Analysis}
We analyze the stopping behavior of FETERS under limited labeled data from two perspectives: consistency between the 5-shot and full-shot settings, and the resulting accuracy–earliness trade-off. These analyses examine whether the selected dataset-level ratios remain reasonably concentrated and whether their stopping behavior leads to an effective balance between prediction accuracy and earliness.

\begin{figure}[!t]
\centering
\includegraphics[width=0.95\columnwidth]{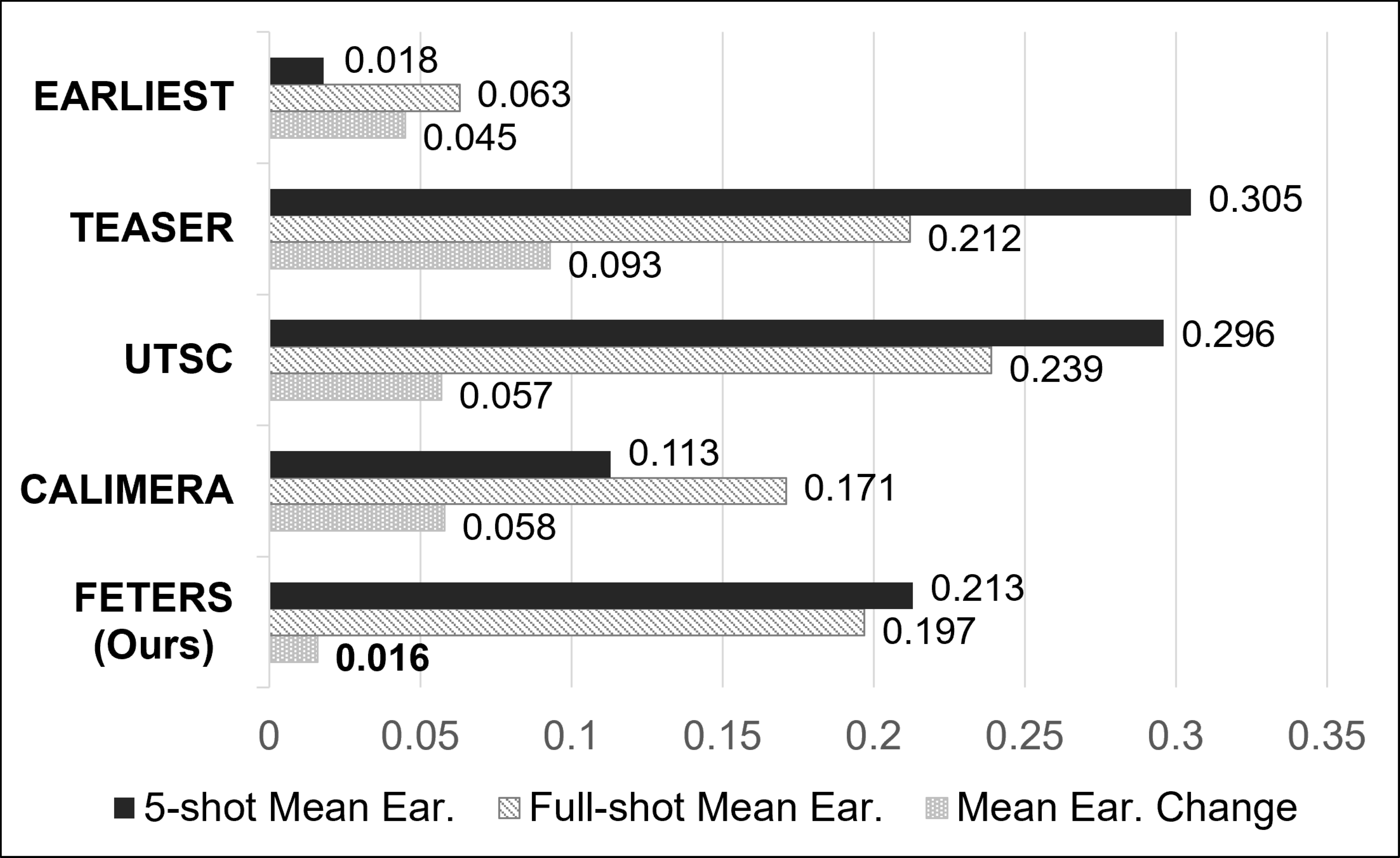}
\caption{Comparison of mean earliness and earliness change between the 5-shot and full-shot settings under the same parameter configuration.}
\label{fig:earchange}
\end{figure}

\noindent
\textbf{Stopping Consistency across Settings}\quad For a given dataset, the 5-shot and full-shot settings share the same underlying temporal characteristics, although the amount of labeled data available for ratio estimation differs substantially. We therefore use the full-shot result as a descriptive reference and define the earliness shift as the absolute difference between the full-shot and 5-shot earliness values. A smaller shift indicates that reducing the support set size results in a smaller change in the selected stopping point. As shown in Figure~\ref{fig:earchange}, FETERS achieves the smallest mean earliness change of 0.016 among the evaluated methods, indicating that its average stopping location changes only slightly when supervision is reduced from the full-shot to the 5-shot setting. Because the two settings use different amounts of labeled data for classifier fitting and ratio selection, this comparison characterizes cross-setting consistency rather than invariance to support-set sampling.

\noindent
\textbf{Effective Accuracy--Earliness Trade-off}\quad Consistency alone does not guarantee effective early classification. A method may produce similar stopping points across settings simply by consistently stopping too early or too late. Under limited supervision, observing a larger portion of the time series may improve accuracy, but the improvement should be sufficient to justify the additional delay. Therefore, we compare the average HM obtained under the same parameter configuration in the 5-shot and full-shot settings (Table~\ref{tab:hm_5shot_fullshot} AVG Mean. 5-shot vs full-shot). FETERS achieves the highest average HM in this comparison, indicating that its relatively small cross-setting earliness shift is accompanied by an effective accuracy–earliness trade-off. Together, Table~\ref{tab:hm_5shot_fullshot} and Figure~\ref{fig:earchange} indicate that FETERS maintains relatively consistent stopping behavior across supervision settings while preserving an effective accuracy--earliness trade-off.

\subsection{Ablation Study}
We evaluate the two components of the hybrid representation by removing either the frozen Chronos features or the dataset-fitted Rocket features. As shown in Table~\ref{tab:ablation}, both ablated variants produce more losses than wins in accuracy and HM under the 5-shot and full-shot settings, supporting the use of the combined representation. Removing Rocket (w/o Roc.) causes the larger degradation. In the 5-shot setting, the variant without Rocket underperforms the complete model on 58 of 69 datasets in accuracy and 62 datasets in HM; the corresponding numbers in the full-shot setting are 59 and 61. This indicates that dataset-fitted Rocket features provide the primary task-specific representation for prefix classification. Removing Chronos (w/o Chr.) results in smaller but still consistent degradation: the complete model achieves 9 and 8 more accuracy wins than losses in the 5-shot and full-shot settings, respectively, while the corresponding HM margins are 17 and 11.

These results suggest that frozen Chronos representations provide complementary temporal information beyond Rocket features and that combining the two feature sources generally produces a more effective accuracy–earliness trade-off. 

\begin{table}[t]
\centering
\setlength{\tabcolsep}{1mm}
\begin{tabular}{llrrr}
\toprule
\multirow{2.5}{*}{Setting}
& \multirow{2.5}{*}{Variant}
& \multicolumn{1}{c}{Acc.}
& \multicolumn{1}{c}{Ear.} 
& \multicolumn{1}{c}{HM} \\
\cmidrule(lr){3-5}
& & W / T / L & W / T / L & W / T / L \\
\midrule
\multirow{2}{*}{5-shot}
& w/o Chr.  & 28 / 4 / \textbf{37} & \textbf{33} / 8 / 28 
& 19 / 14 / \textbf{36} \\
& w/o Roc.  & 11 / 0 / \textbf{58}  & 19 / 1 / \textbf{49}  & 7 / 0 / \textbf{62} \\
\midrule
\multirow{2}{*}{Full-shot}
& w/o Chr.  & 26 / 9 / \textbf{34} & 8 / 46 / \textbf{15} 
& 24 / 10 / \textbf{35} \\
& w/o Roc.  & 7 / 3 / \textbf{59} & 18 / 18 / \textbf{33} 
& 5 / 3 / \textbf{61} \\
\hline
\end{tabular}
\caption{Ablation study of the hybrid feature representation. W/T/L denotes the number of datasets on which the corresponding ablated variant outperforms, ties with, or underperforms the complete FETERS model.}
\label{tab:ablation}
\end{table}

\subsection{Penalty Parameter and Ratio Selection}
We examine the effect of the penalty parameter $p$ under both supervision settings. As shown in Table~\ref{tab:penalty_parameter}, increasing $p$ generally permits later stopping and yields higher accuracy, whereas removing the penalty results in substantially later predictions and a large decrease in HM. This confirms that the penalty term is necessary for preventing ratio selection from being driven solely by accuracy. The similar results obtained with $p=1.0$, and $p=1.5$ also indicate limited sensitivity to nearby values.

\begin{table}[t]
\centering
\begin{tabular}{llccc}
\toprule
Setting & Penalty & Acc. $\uparrow$ & Ear. $\downarrow$ & HM $\uparrow$ \\
\midrule
\multirow{5}{*}{5-shot}
& $p=0.6$ & 0.598 & 0.165 & 0.677 \\
& $p=1.0$ & 0.618 & 0.195 & 0.680 \\
& $p=1.5$ & 0.633 & 0.226 & 0.678 \\
& $p=2.0$ & 0.644 & 0.253 & 0.672 \\
& w/o penalty & 0.721 & 0.594 & 0.436 \\
\midrule
\multirow{5}{*}{Full-shot}
& $p=0.6$ & 0.688 & 0.151 & 0.747 \\
& $p=1.0$ & 0.709 & 0.180 & 0.748 \\
& $p=1.5$ & 0.738 & 0.217 & 0.746 \\
& $p=2.0$ & 0.749 & 0.254 & 0.738 \\
& w/o penalty & 0.843 & 0.704 & 0.372 \\
\bottomrule
\end{tabular}
\caption{Ablation and sensitivity analysis of the penalty-based reward function under different values of $p$.}
\label{tab:penalty_parameter}
\end{table}

Table~\ref{tab:fix_ratio} further compares FETERS with representative globally fixed ratios near the selected earliness mean. FETERS achieves the highest average HM under both settings. In particular, compared with $r=0.20$, which provides similar average earliness, FETERS obtains substantially higher accuracy and HM. These results support selecting a dataset-specific ratio rather than applying a common observation ratio to all datasets.

\begin{table}[t]
\centering
\begin{tabular}{llccc}
\toprule
Setting & Ratio & Acc. $\uparrow$ & Ear. $\downarrow$ & HM $\uparrow$ \\
\midrule
\multirow{7}{*}{5-shot}
& $r=0.05$ & 0.464 & 0.050 & 0.586 \\
& $r=0.10$ & 0.509 & 0.098 & 0.616 \\
& $r=0.15$ & 0.544 & 0.147 & 0.631 \\
& $r=0.20$ & 0.577 & 0.198 & 0.642 \\
& $r=0.25$ & 0.606 & 0.249 & 0.645 \\
& $r=0.30$ & 0.631 & 0.298 & 0.643 \\
& \textbf{FETERS} & 0.628 & 0.213 & \textbf{0.679} \\
\midrule
\multirow{7}{*}{Full-shot}
& $r=0.05$ & 0.540 & 0.050 & 0.651 \\
& $r=0.10$ & 0.596 & 0.098 & 0.685 \\
& $r=0.15$ & 0.639 & 0.147 & 0.702 \\
& $r=0.20$ & 0.677 & 0.198 & 0.710 \\
& $r=0.25$ & 0.712 & 0.249 & 0.710 \\
& $r=0.30$ & 0.736 & 0.298 & 0.702 \\
& \textbf{FETERS} & 0.725 & 0.197 & \textbf{0.749} \\
\bottomrule
\end{tabular}
\caption{Comparison between FETERS and globally fixed ratios from 0.05 to 0.30, focusing on the early-decision region where the accuracy--earliness trade-off is most relevant.}
\label{tab:fix_ratio}
\end{table}

\section{Conclusion}
We presented FETERS, a few-shot ETSC framework that selects a dataset-level stopping ratio through support-set class-wise LOO evaluation and combines dataset-fitted Rocket features with frozen Chronos representations for prefix classification. Experiments on 69 datasets across 14 domains show that FETERS achieves the best average HM and HM rank among the evaluated methods in the 5-shot setting while remaining competitive under full-shot supervision. These results demonstrate the effectiveness of FETERS in managing the accuracy–earliness trade-off under limited labeled data.


\bibliography{reference}


\end{document}